\documentclass[a4paper, 12pt, twoside]{article}
\input{JADTstyle.cls}
\usepackage{array}
\usepackage{booktabs}
\usepackage{float}
\usepackage{caption}
\usepackage{microtype}
\usepackage{graphicx}
\usepackage{tabularx}  
\usepackage{ragged2e}  
\usepackage[none]{hyphenat}
\usepackage{dirtytalk}
\usepackage{csquotes}

\begin{document}

{\centering
\fontsize{18}{22}\selectfont
Economic Transformation and Cultural Change: Evidence from Two Centuries of French Drama\par
\vspace{1.4em}

\fontsize{14}{18}\selectfont
\makebox[\textwidth][c]{%
  \makebox[0.38\textwidth][r]{Thiago D. Oliveira\textsuperscript{1}}%
  \hspace{1.0cm}%
  \makebox[0.38\textwidth][l]{Luccas A. Attilio\textsuperscript{2}}%
}\par
\vspace{0.7em}
Marwil J. Davila-Fernandez\textsuperscript{3}\par

\vspace{1em}
\fontsize{12}{16}\selectfont
\textsuperscript{1}\textit{Centre for Digital Humanities, University of Tartu}\par
\textsuperscript{2}\textit{Federal University of Ouro Preto}\par
\textsuperscript{3}\textit{Colorado State University}\par
}

\thispagestyle{plain}

\selectlanguage{english} 

\setlength{\parindent}{\z@}
\setlength{\baselineskip}{11pt}


\vspace*{2pt}
\abstract*{Abstract}

\begin{footnotesize}{ 
How do large-scale economic transformations shape cultural production? We address this question by combining computational linguistics, econometrics, and formal modelling, using French drama as a well-documented empirical laboratory. Applying latent Dirichlet allocation to a corpus of 1,215 theatrical texts published between 1700 and 1900, we show that aristocratic discourse centred on sovereignty and political authority was gradually displaced by bourgeois and household economic themes as French capitalism developed. Bayesian vector autoregressive models with max-share shock identification suggest a temporal shift in the literary response to economic shocks: bourgeois everyday-life themes reacted to GDP shocks in the eighteenth century, whereas household-economic concerns became responsive only after 1820, amid accelerating industrialisation. A discrete-choice model shows that peer effects among authors and sensitivity to prevailing economic conditions can jointly account for these dynamics. Monte Carlo simulations reproduce the observed historical trajectory with reasonable fidelity. These findings offer a quantitative framework for understanding how economic transformations propagate into cultural production through identifiable social mechanisms, contributing to the study of cultural evolution and the long-run relationship between institutions and literary discourse.

}\end{footnotesize}

\vspace*{14pt}\begin{footnotesize}{\textbf{Keywords:} Political Economy, Cultural History, Computational Linguistics, BVAR Model, Discrete Choice Theory.}\end{footnotesize}


\setlength{\parskip}{6pt}
\setlength{\baselineskip}{14pt}

\section{Introduction}

The relationship between economic transformation and cultural production is among the oldest questions in the social sciences. Yet, it has proved remarkably resistant to systematic empirical investigation. Cultural change happens over centuries, making it hard to measure with the precision required for causal analysis. Recent advances in computational linguistics and econometric methods now make it possible to approach this question rigorously \citep{michel2011quantitative}.

We investigate how two of the most consequential transformations in modern history, the French Revolution and the industrialisation of France, reshaped literary production, examining them from a \textit{longue durée} perspective \citep{braudel1958histoire}. Using a corpus of 1,215 French plays published between 1700 and 1900, we trace the long-run evolution of dramatic themes against the shifting political-economic landscape of France \citep{hobsbawmrevolution, hobsbawmcapital}. Epistemologically, we treat literary production as embedded in political-economic relations \citep{granovetter1985economic}, yet not fully determined by material conditions, since individuals simultaneously shape and are shaped by institutions \citep{agassi1975institutional}. The interplay between drama and the political-economic world is thus an instance of the longstanding debate regarding the relationship between agent and structure \citep{oliveira2018nature}. We concur with \cite{mandelbaum1955societal} that individual action cannot be understood without reference to \say{societal facts}, as \say{the actual behaviour of specific individuals towards one another is unintelligible unless one views their behaviour in terms of their status and roles, and the concepts of status and role are devoid of meaning, unless one interprets them in terms of the organisation of the society to which the individuals belong} (pp. 307-308).

Literary production is deeply rooted in political-economic relations, and is best approached as a \say{total social fact} \citep{mauss1925gift}. France between 1700 and 1900 offers an unusually well-documented laboratory for this investigation. The political rupture of 1789 and the subsequent acceleration of industrialisation from the 1820s onward provide sharp disruptions against which cultural change can be measured. Drama is especially well suited to tracing this embeddedness. It is a publicly performed and commercially mediated art form at the intersection of aesthetic production and social demand. These characteristics make it a privileged site where the shifting preoccupations of a society become textually legible at scale.

Topic modelling has been increasingly used within computational literary studies. \cite{schoch2021topic} investigates whether topics obtained through latent Dirichlet allocation (LDA) correspond to genres and subgenres of French drama from 1630 to 1789. \cite{min2019modeling} use Non-Negative Matrix Factorisation (NMF), sentiment analysis, and network analysis to study Victor Hugo's \textit{Les Misérables}. \cite{dahllof2019faces} combine gender studies and Swedish literary history, using topic modelling for a gendered thematic analysis of Swedish prose fiction. \cite{ginn2024historia} compare LDA, NMF and Bidirectional Encoder Representations from Transformers (BERT) type models to examine Roman literature, while \cite{martinelli2024exploring} apply neural topic modelling to Classical Latin literature. Our paper builds on this tradition by connecting computational literary analysis to econometric methods and formal economic modelling, allowing us to move from description to causal inference and theoretical interpretation within a single unified framework.

We show that the thematic structure of French drama underwent a profound transformation coinciding with the French Revolution and the industrialisation of the country. Aristocratic discourse centred on sovereignty, virtue, and political authority was systematically displaced by bourgeois and household economic themes as capitalist social relations deepened. We contribute to computational literary studies, cultural history, and political economy by showing how shifts in literary production respond to long-term economic and institutional change, and by identifying the social mechanisms through which this process unfolds.

\section{Data and Methods}

Our data are 1215 French plays published between 1700 and 1900. The corpus was built using the French-language corpora made available by the DraCor project \citep{fischer2019programmable}, and it was retrieved via their API (\url{https://dracor.org/doc/api}). Documents were processed using spaCy's French language model and normalised through lemmatisation because French is a highly inflected language. Preprocessing used a part-of-speech filter to retain only nouns, proper nouns, verbs, and adjectives. French stop words were removed, as well as highly frequent auxiliary verbs (e.g. être, avoir, faire, dire) and address terms (e.g. monsieur, madame), since they are essentially grammatical words without lexical meaning. In order to assess the changing nature of French drama in the 18th and 19th centuries, we use LDA and Jensen-Shannon divergence (JSD). 

LDA is a machine learning method widely used in topic modelling to identify the latent semantic structure of texts. It is a generative probabilistic model that represents documents as probability distributions over latent topics, while topics are treated as probability distributions over the vocabulary (\citealp{blei2003latent}; \citealp{steyvers2007probabilistic}; \citealp{blei2012probabilistic}). Instead of reading \textit{between} the lines to uncover authors’ purposes, LDA shifts the analysis to reading \textit{above} the lines, such that the topical structure of the texts is derived from the entire constellation of documents to which they belong. It is a powerful instrument in approaching the coevolution of literature and economics from a \say{distant reading} perspective, rather than focusing on the exegesis of a few canonical plays \citep{moretti2013distant}. The method is particularly well suited for exploring questions that bridge literary studies and socioeconomic perspectives on culture \citep{dimaggio2013exploiting}.

One of the crucial issues with LDA is the selection of the number of topics, a hyperparameter of the model which must be decided a priori. The optimal number of topics depends on the characteristics of the corpus, on the number of tokens per document, and on the level of granularity adequate to the research question (\citealp{rhody2012topic}; \citealp{jockers2013macroanalysis}; \citealp{sbalchiero2020topic}). Although there are numerous studies on the detection and selection of the number of topics (\citealp{griffiths2004finding}; \citealp{newman2010automatic}; \citealp{roder2015exploring}; \citealp{gan2021selection}), the optimal number of topics obtained via metrics such as perplexity and topic coherence is hardly a good benchmark,  since the appropriate number of topics ultimately depends on topic interpretability and on the level of granularity required by the problem at hand \citep{wehrheim2019economic}. Thus, while LDA is an unsupervised machine learning technique, historical analysis and a theoretically informed understanding of the object of study remain essential in evaluating the coherence of topics and in selecting the number of topics \citep{mohr2013introduction}.

Our choices regarding preprocessing and the number of topics were based on topic interpretability and coherence, following experiments with different Part-of-Speech (POS) strategies and different numbers of topics. Since our main goal is to trace the rise of economic discourse in French drama, a 10-topic model proved sufficient for yielding topics with a strong presence of economic discourse and themes of bourgeois everyday life, as well as topics reminiscent of Greek tragedy and the aristocratic zeitgeist. As a robustness check, we also used NMF for the topic modelling analysis, which yielded similar results, as reported in the Appendix.

In order to identify periods of rapid topical change in French drama, we use Jensen-Shannon divergence (\citealp{lin2002divergence}; \citealp{nielbo2023pandemic}). JSD measures how much the topic distribution of documents changes between 2 consecutive years. It is defined by

\[\mathrm{JSD}(P \mid Q)=\frac{1}{2} \mathrm{D}(P \mid M)+\frac{1}{2} \mathrm{D}(Q\mid M)\]

where $P$ and $Q$ are the two probability distributions we want to compare, \( M = \frac{1}{2}(P + Q) \) and \( D \) is the Kullback--Leibler divergence: 

\[D(P \mid Q) = \sum_{i : P_i > 0} P_i \log_2 \frac{P_i}{Q_i}\]

Lastly, we use multidimensional scaling to construct a semantic map based on the cosine distances between documents’ topic vectors. Three labels were added on the map, corresponding to three topics of particular interest to our argument, which we further explore throughout the paper. The labels were placed at the centroid of the five representative works listed in Table 1, i.e. the five documents with the highest topic proportions for each topic of interest.

\section{Thematic Transformation in French Drama, 1700–1900}

Table 1 reports the average topic prevalence for the periods 1700–1800 and 1800–1900, along with the top 15 words and five representative works per topic. Comparing topic prevalence across time windows gives us insight into which themes lost or gained prominence after the French Revolution and the industrialisation of the country. Three topics stand out as highly prevalent in one of the windows, but not the other, namely topics 4, 8 and 9. Whereas the importance of topic 8 falls from 0.25 to 0.09, that of topics 4 and 9 increase to 0.23 and 0.22, respectively. Since we are mostly interested in diachronic linguistic change, in what follows we focus our analysis on these three topics. For the ease of interpretation, we label them as \textit{Bourgeois Life} (topic 4), \textit{Aristocratic Life} (topic 8) and \textit{Household Economics} (topic 9).
   
\begin{table}[tbp]
\centering
\begingroup
\scriptsize
\setlength{\tabcolsep}{3pt}
\renewcommand{\arraystretch}{1.05}
\sloppy
\begin{tabularx}{\textwidth}{@{}
>{\centering\arraybackslash}p{0.7cm}
>{\centering\arraybackslash}p{2.4cm}
>{\centering\arraybackslash}p{2.4cm}
>{\RaggedRight\arraybackslash}p{4.0cm}
>{\RaggedRight\arraybackslash}X@{}}
Topic & \shortstack{Prevalence$_{1700-1800}$}
      & \shortstack{Prevalence$_{1800-1900}$}
      & Top Words (15)
      & Representative Works \\
\midrule

0 & 0.07 & 0.06 &
French, author, France, art, talent, taste, work, Paris, public, genius, nature, opera, subject, truth, state &
Diderot: \textit{Entretien entre d'Alembert et Diderot}\newline
Anonymous: \textit{Ouverture de la Séance}\newline
Sacy: \textit{L'Ile Déserte ou Le Naufrage}\newline
Diderot: \textit{Rêve de d'Alembert}\newline
Sacy: \textit{Le Testament} \\
\addlinespace[4pt]

1 & 0.13 & 0.12 &
mother, secret, sentiment, brother, son, uncle, sister, letter, friendship, care, soul, hope, deceive, tender, tenderness &
Genlis: \textit{La Mère Rivale, Comédie}\newline
Colleville: \textit{Sophie et Derville}\newline
Carmontelle: \textit{Les Faux Indifférents}\newline
Carmontelle: \textit{Le Faux Empoisonnement}\newline
Graffigny: \textit{Cénie} \\
\addlinespace[4pt]

2 & 0.02 & 0.06 &
Molière, play, theatre, play, comedy, stage, role, actor, comedian, author, poet, king, gift, talent, laugh &
Jouy: \textit{Les Bancs de la Promenade}\newline
Diderot: \textit{Paradoxe sur le Comédien}\newline
Sacy: \textit{La Modestie}\newline
Sacy: \textit{La Répétition}\newline
Chalmeton: \textit{A Jean Racine} \\
\addlinespace[4pt]

3 & 0.10 & 0.06 &
again, good, husband, affair, care, fool, wise, marriage, turn, speech, madman, devil, laugh, law, old &
Voltaire: \textit{Le Dépostaire}\newline
Voltaire: \textit{L'Enfant Prodigue}\newline
Regnard: \textit{Démocrite}\newline
Regnard: \textit{Les Ménechmes, ou Les Jumeaux}\newline
Barbier: \textit{Le Faucon} \\
\addlinespace[4pt]

4 & 0.02 & 0.23 &
poor, old, night, evening, king, mother, flower, death, dream, sun, voice, fear, soul, arm, memory &
Beissier: \textit{L'Oiseau Bleu, Saynète}\newline
Adenis: \textit{Le Nouveau Né}\newline
Monselet: \textit{Par la Poste}\newline
Crossonnois: \textit{Un Monsieur qui ne Veut Plus Fumer}\newline
Beissier: \textit{La Nuit de Noël, Comédie} \\
\addlinespace[4pt]

5 & 0.15 & 0.04 &
affair, marry, marriage, Lisette, husband, knight, wed, mistress, mother, farewell, anger, sister, honest, marquis, truth &
Dancourt: \textit{Colin-Maillard}\newline
Marivaux: \textit{Le Dénouement Imprévu}\newline
Marivaux: \textit{La Commère}\newline
Audiffret: \textit{L'Epreuve}\newline
Marivaux: \textit{La Double Inconstance} \\
\addlinespace[4pt]

6 & 0.10 & 0.04 &
lover, tender, sweet, fire, kind, ardor, charming, charm, object, flame, beauty, Leander, gentle, desire, soul &
Voisenon: \textit{Hilas et Zélis}\newline
Saint-Gilles Lenfant: \textit{La Feinte Heureuse}\newline
La Motte: \textit{Le Carnaval et la Folie}\newline
Favart: \textit{L'Amant Déguisé}\newline
Hénault: \textit{Le Temple des Chimères} \\
\addlinespace[4pt]

7 & 0.03 & 0.08 &
death, son, land, mother, city, word, dwelling, king, foreigner, kill, brother, blood, old man, enemy, sea &
Anonymous: \textit{Oedipe à Colone}\newline
Crébillon: \textit{Electre}\newline
Anonymous: \textit{Oedipe Roi}\newline
La Harpe: \textit{Philoctète}\newline
Beissier: \textit{Antigone} \\
\addlinespace[4pt]

8 & 0.25 & 0.09 &
king, blood, son, death, fate, virtue, law, people, crime, glory, cruel, care, fury, again, horror &
Chamfort: \textit{Mustapha et Zéangir}\newline
Crébillon: \textit{Pyrrhus}\newline
Voltaire: \textit{Saül}\newline
Chabanon: \textit{Eudoxie}\newline
Crébillon: \textit{Atrée et Thyeste} \\
\addlinespace[4pt]

9 & 0.13 & 0.22 &
money, devil, affair, poor, pay, husband, house, franc, rogue, eat, lady, letter, father, boy, drink &
Archambault: \textit{Janot}\newline
Archambault: \textit{Janot chez le Dégraisseur}\newline
Renard: \textit{La Demande}\newline
Gueullette: \textit{Le Chapeau de Fortunatus}\newline
Carmontelle: \textit{Les Voyageurs} \\
\end{tabularx}
\caption{\small Topic Prevalence 1700–1800 vs. 1800–1900, 15 Top Words and 5 Representative Works.}
\endgroup
\end{table}

Topics were labelled by manually inspecting the plot and main themes of plays with prevalence above 0.5 for each of the three topics. Considering topic 4, for instance, there are 69 documents satisfying the 0.5 threshold. While the top 15 words shown in Table 1 do not clearly indicate what topic 4 is about, an inspection of these 69 documents reveals that they are mostly sentimental and social comedies dealing with ordinary aspects of everyday life. We thus chose the label \textit{Bourgeois Life} for this topic to emphasise that its plays are more preoccupied with urban life and the emergent middle class than with courtly life and the aristocracy. The label aims simply to distinguish the emerging middle class from the then-dominant aristocratic class. 
Topic 8, or \textit{Aristocratic Life}, focuses on questions of virtue, justice, and political authority, which are reminiscent of the absolutist state. Its top words include \textit{king}, \textit{virtue}, \textit{law}, \textit{people}, \textit{crime}. This topic thus touches on issues regarding the political organisation of the state, in modern parlance, that were central to political philosophers from Plato's \textit{Republic} and Aristotle's \textit{Politics} to Machiavelli's \textit{Prince} and Hobbes's \textit{Leviathan}. Finally, we chose \textit{Household Economics} as a label for topic 9 because it has a clear material dimension, as can be seen by the presence of \textit{money}, \textit{poor}, \textit{pay}, \textit{husband}, \textit{house} and \textit{franc} among its top 15 words. Therefore, both topics 4 and 9 deal with the emerging middle class, which renders them quite different from topic 8. The two topics differ, however, in that topic 9 has a strong economic component, whereas topic 4 deals with the social life of the nascent bourgeois class. 

Fig. 1 shows the evolution of these three topics vis-à-vis French GDP.\footnote{We use real GDP per capita in 2011 US dollars. The data were retrieved from the Maddison Project and are available from \url{https://www.rug.nl/ggdc/historicaldevelopment/maddison/}} The decreasing importance of \textit{Aristocratic Life} since the 1750s seems to reflect the rise of anti-monarchic sentiment during this period, inspired by Montesquieu's \textit{The Spirit of Laws} (1748), Rousseau's \textit{The Social Contract} (1762), and Diderot \& D'Alembert's \textit{Encyclopédie} (1751-1772). On the other hand, there is a striking similarity between the evolution of GDP and \textit{Household Economics}. Both series grew very slowly during the 18th century (although \textit{Household Economics} oscillated more than GDP), and both series took off between 1825 and 1850. GDP growth rates increased considerably after 1825 as France initiated its industrialisation, with the first railway being completed in 1827 (Saint-Étienne to Andrézieux). The series \textit{Bourgeois Life} also presents a behaviour quite similar to GDP throughout most of the 18th century, although not as much in the 19th. Notice that both series grow in tandem until roughly the beginning of the French Revolution, when the slope of \textit{Bourgeois Life} increases considerably. After 1850, however, the series becomes essentially flat, while GDP continues to increase steadily. In the next section, we use econometric methods to investigate the impact of GDP shocks on these three topics, or lack thereof.

\begin{figure}[tbp]
\centering
\includegraphics[scale=0.5]{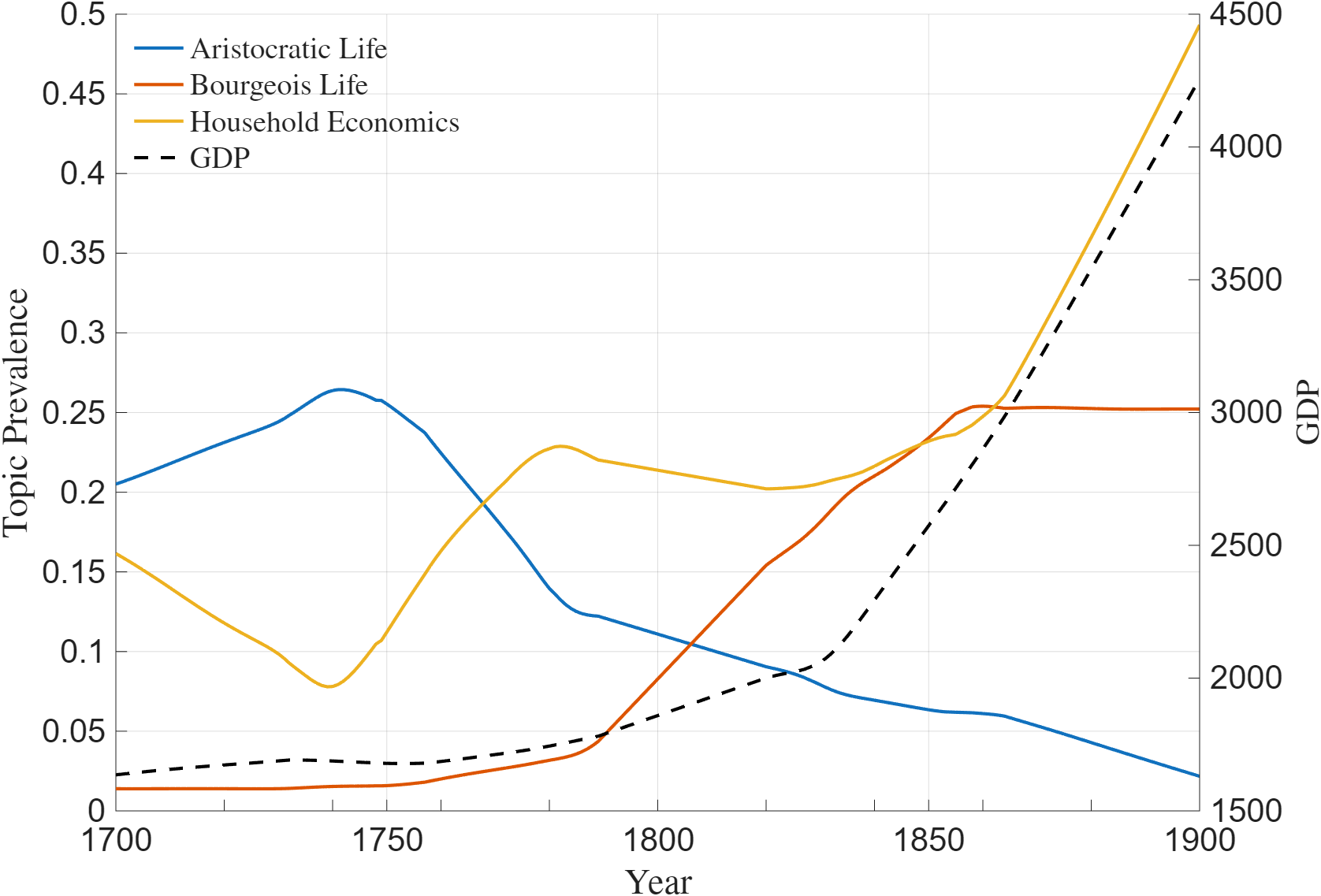}
	\caption{LOWESS-smoothed series of Topic Prevalence vs Real GDP per capita in 2011 USD.}
\end{figure}

The development of French capitalism thus marks a shift in literary production from topics that reflect on the political authority of rulers to more mundane themes, reflecting the material reality of the nascent bourgeoisie. Continued growth and prosperity throughout the rest of the century, coupled with the proclamation of the Third Republic in 1870, completely shifted the politico-economic milieu in which authors and readers were socially embedded. Such changes led to the commercialisation of theatre, to increased literacy rates, and to an expansion of middle-class audiences.  The growing importance of \textit{Bourgeois Life} and \textit{Household Economics} maps rather neatly into Hobsbawm's periodisation in \textit{The Age of Revolution: 1789–1848} and \textit{The Age of Capital: 1848–1875} \citep{hobsbawmrevolution,hobsbawmcapital}. While \textit{Bourgeois Life} grows in importance during the Age of Revolution, it flattens during the Age of Capital. \textit{Household Economics}, on the other hand, would only increase in importance in the 1850s, precisely when Hobsbawm identifies the consolidation of capitalism. As French capitalism developed and a bourgeois class emerged in the country, the nature of French drama changed drastically. 

Jensen-Shannon Divergence analysis shows that the textual profile of French drama changed sharply in the first half of the nineteenth century. The left panel of Fig. 2 reports this finding. The peak in the JSD series reflects the concomitant decrease in prevalence of \textit{Aristocratic Life} and increase in prevalence of \textit{Bourgeois Life} and \textit{Household Economics} between 1800 and 1850. The French Revolution and the industrialisation of the country thus led to a revolution in French drama, which shifted toward socio-economic themes concerning the middle class and away from discussions of sovereignty, virtue and violence. This pattern is also visible in the semantic map shown in the right panel of Fig. 2, where each node in the map corresponds to a play, and plays located close to the labels are more strongly associated with those topics. Accordingly, plays from the second half of the nineteenth century cluster in the lower-left region, indicating a shared textual profile that distinguishes them from eighteenth-century plays. 

\begin{figure}[tbp]
\centering
\begin{minipage}[t]{0.49\textwidth}
  \centering
  \includegraphics[scale=0.33]{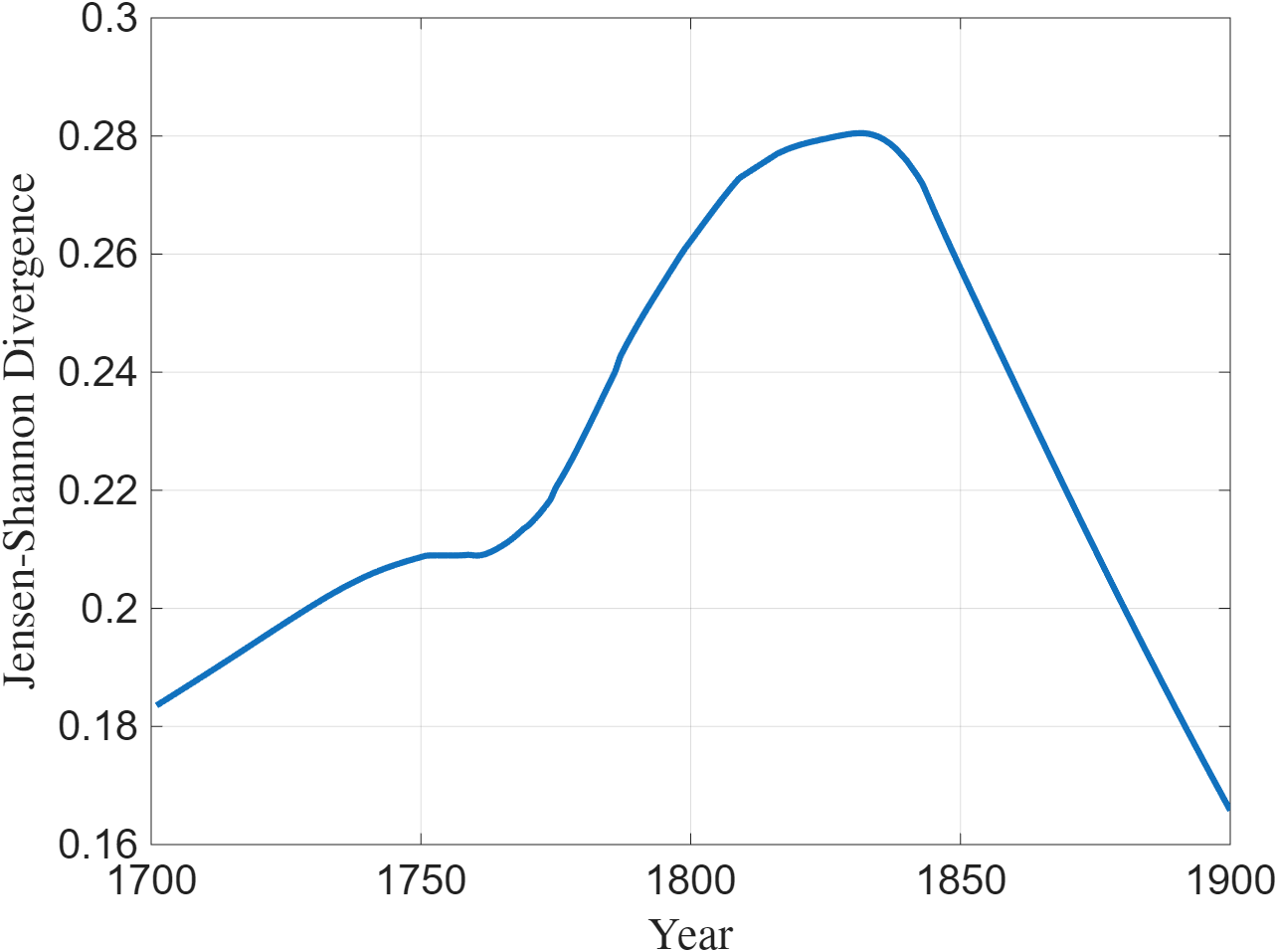}
\end{minipage}\hfill
\begin{minipage}[t]{0.49\textwidth}
  \centering
  \includegraphics[scale=0.3]{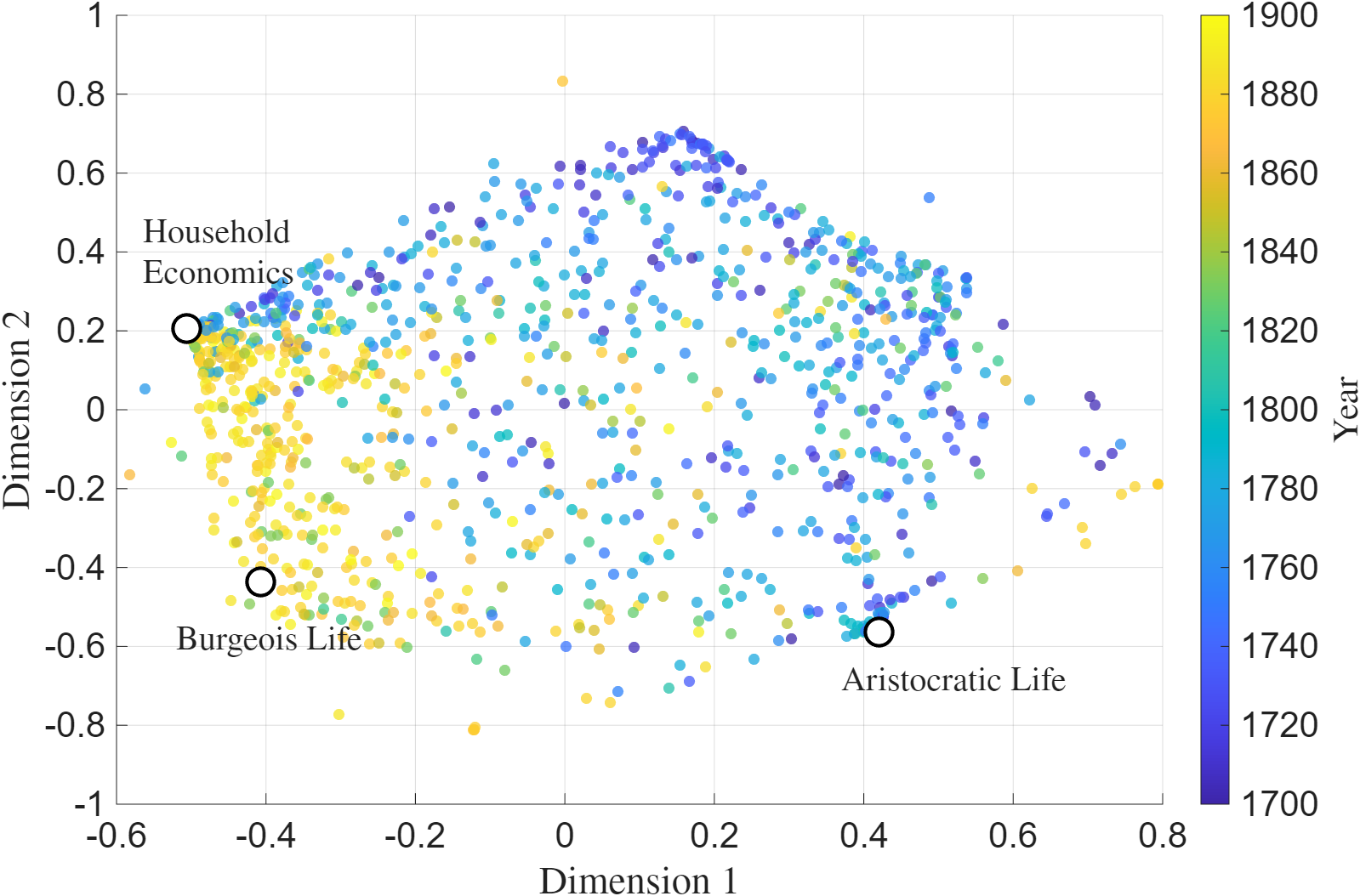}
\end{minipage}
\caption{Jensen-Shannon Divergence (left) and semantic map (right). The map was constructed using Multidimensional Scaling (MDS) of cosine distances between LDA topic-distribution vectors. Labels indicate the centroids of the five representative documents listed in Table 1.}
\end{figure}

\section{The Political Economy of Literary Production: BVAR Analysis}

Establishing that literary themes and GDP moved together is suggestive, but does not tell us whether economic growth drove cultural change. To address this, we use a Bayesian Vector Autoregressive (BVAR) model that identifies a structural shock from a linear combination of reduced-form shocks based on indices capturing the prevalence of topics and GDP per capita. The estimates allow us to assess how the three topics discussed in the previous section respond to changes induced by this structural shock. The derivation of the BVAR model is presented in Appendix A2.1. 

We use the max-share identification strategy \citep{angeletos2020,chahrour2024exchange}. This approach overcomes a major limitation of classical identification schemes, such as the Cholesky decomposition, by avoiding the need to impose a recursive ordering of variables from most to least exogenous. The max-share identification is a data-driven approach where shocks are determined by exploiting the information contained in the full VAR system. It recovers the structural shock that maximises the Forecast Error Variance Decomposition (FEVD) share of a target variable, GDP in our case, at a given horizon.

The macroeconometric literature has used BVARs under max-share shock identification to analyse the main drivers of real gross national product for the US between 1964-2001 \citep{uhlig2003moves}, to explore the impact of news shocks on the fluctuations of future total factor productivity \citep{barsky2011news}, to replicate stylised facts about business cycles in the US between 1955-2017 \citep{angeletos2020}, and to analyse the relationship between US productivity and the exchange rate between the US dollar and G7 currencies \citep{chahrour2024exchange}. These studies illustrate one of the main advantages of our identification strategy; since macroeconomic variables are influenced by multiple factors, the max-share approach identifies the main drivers of variation in a given variable, allowing the data to speak more freely, with fewer a priori restrictions.

The BVAR adopts the Minnesota prior, which places higher weight on each variable's own lags, especially the first lag, while shrinking coefficients on longer lags and cross-variable effects \citep{angeletos2020,chahrour2024exchange}. The posterior distribution was obtained from 50,000 Gibbs-sampling draws, following \citet{chahrour2024exchange}. Based on the Akaike and Bayesian information criteria, we adopt two lags in the base model. The shocks correspond to one standard deviation, and impulse responses are reported with 68\% confidence intervals. The empirical model uses the logarithms of GDP per capita and of the constructed indices: \textit{Household Economics}, \textit{Bourgeois Life}, and \textit{Aristocratic Life}. The sample covers the period from 1700 to 1900, using annual data.

The analysis is divided into two sub-periods, 1700–1789 and 1820–1900, for two reasons. First, the corpus contains very few plays published during the Revolutionary and Napoleonic periods, reflecting the well-documented disruption of French cultural institutions between 1789 and the early Restoration. Theatres were closed, censored, or repurposed, and dramatic output fell sharply before recovering under the July Monarchy. Second, the two sub-periods correspond to historically distinct politico-economic regimes. They separate the pre-Revolutionary years of slow growth and emerging bourgeois culture, and the post-Restoration era of accelerating industrialisation.

\subsection{Baseline Model: 1700–1789}
Fig. 3 displays a structural GDP shock and the responses of the variables for the period from 1700 to 1789. Although \textit{Household Economics} and \textit{Aristocratic Life} increase, their estimates are not statistically significant. \textit{Bourgeois Life} increases during all four years following the shock, with statistically significant values. The estimates indicate that the time series representing bourgeois everyday life themes in French drama responded strongly to changes in economic conditions. The GDP shock explains 64.3\% of its short-term fluctuations and 94.1\% in the long term, as indicated in Table 2, which shows the influence of the GDP shock on all variables. Hence, although GDP and \textit{Bourgeois Life} grew modestly in the 18th century, the two series followed closely parallel trajectories and our results indicate that literary production slowly adjusted to economic growth.

\begin{figure}[tbp]
\centering
\includegraphics[scale=0.875]{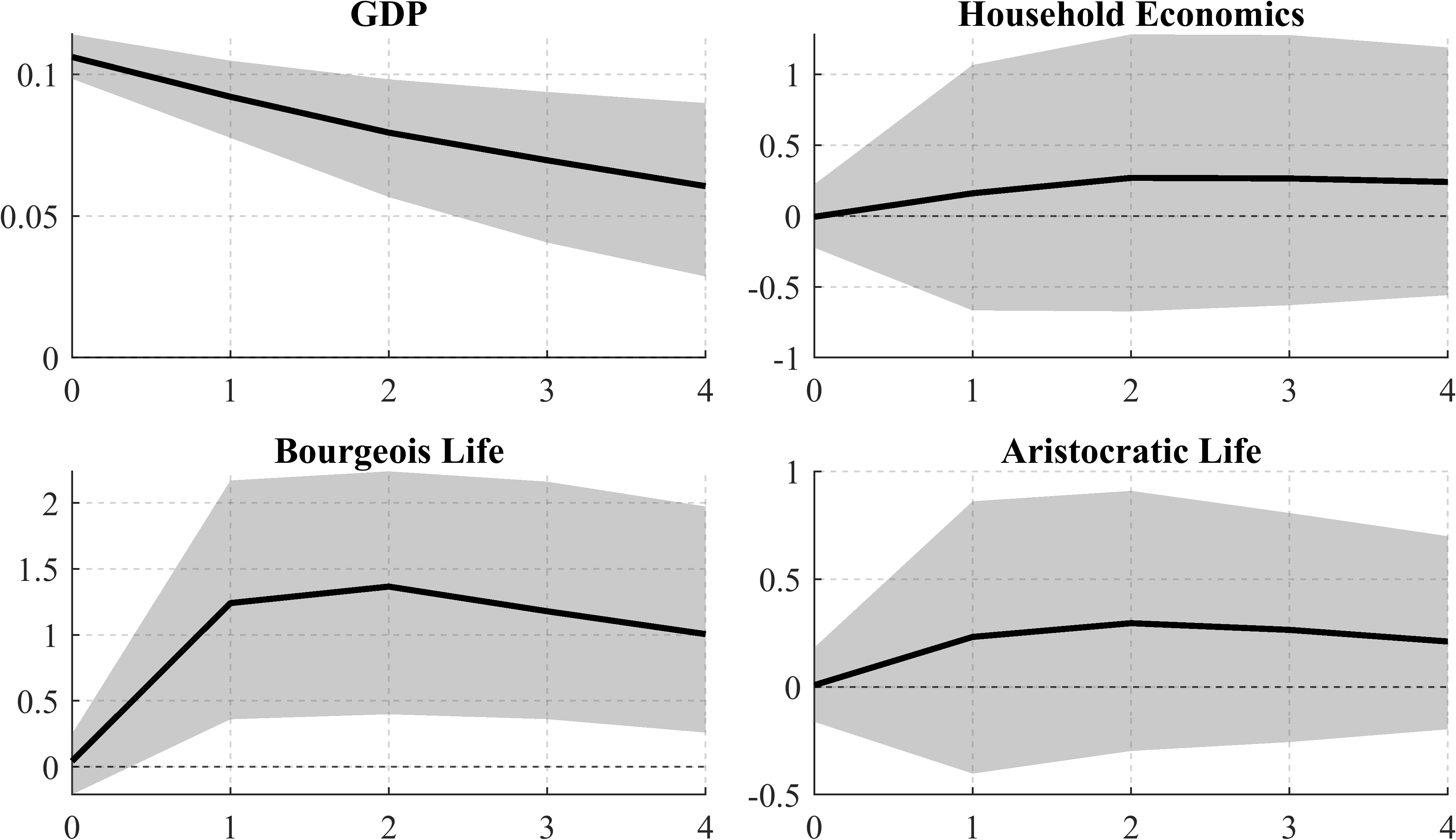}
	\caption{BVAR model, 1700–1789, GDP shock and responses of the variables}
    \caption*{\footnotesize Note: Impulse responses to a one standard deviation structural GDP shock. Solid lines denote posterior mean estimates and shaded areas denote 68\% confidence intervals. Responses whose confidence intervals exclude zero are statistically significant.}
\end{figure}

As reported in Table 2, the shock explains 32.8\% of the fluctuation in \textit{Household Economics}, 64.3\% in \textit{Bourgeois Life}, and 33.7\% in \textit{Aristocratic Life} in the short term. In 20 years, denoted as the long term, the values increase for all variables, especially for \textit{Bourgeois Life}. The strongest influence of the GDP shock is on \textit{Bourgeois Life}. In contrast, \textit{Aristocratic Life} and \textit{Household Economics} display limited responsiveness to economic fluctuations in this period. One interpretation of this result is that economic growth in the 18th century, coupled with the political debates leading up to the French Revolution, provided momentum for a gradual shift within theatrical texts whereby the focus increasingly becomes on the middle class and the business of ordinary life.

\setcounter{table}{1}
\begin{table}[tbp]
\centering
\caption{Impact of the structural shock on variables, 1700–1789}
\resizebox{\textwidth}{!}{
\begin{tabular}{lcccc}
\hline
 & GDP & Household Economics & Bourgeois Life & Aristocratic Life \\
\hline
\textbf{Short term (0--4 years)}  & 100.0 & 32.8 & 64.3 & 33.7 \\
                                   & [100.0, 100.0] & [7.6, 65.3] & [24.9, 82.4] & [8.9, 65.4] \\
\addlinespace
\textbf{Long term (20+ years)}   & 100.0 & 79.0 & 94.1 & 80.6 \\
                                   & [100.0, 100.0] & [19.4, 97.6] & [63.2, 99.4] & [25.2, 98.1] \\
\hline
\end{tabular}
}
\caption*{\footnotesize Notes: Entries report the share of forecast error variance in each variable explained by the structural GDP shock, expressed as percentages. Short and long term correspond to horizons of 0 to 4 and 20 years respectively. Brackets denote 68\% confidence intervals. Higher values indicate greater sensitivity to economic conditions.}
\end{table}

\subsection{Baseline Model: 1820–1900}
Fig. 4 and Table 3 present our estimates for the period 1820-1900. \textit{Aristocratic Life} remains unaffected by GDP shocks. However, the response of \textit{Bourgeois Life} to the GDP shock is no longer statistically different from zero, while \textit{Household Economics} becomes sensitive to economic changes. GDP shocks explain 32.8\% of its variation in the short term and 95.3\% in the long term. Given that French industrialisation would only accelerate in the 1820s, this could explain why economic issues appear to have only entered the stage of French drama in the mid 1850s. Although the topic had always had a non-negligible presence, its average prevalence increased from 0.13 in 1700-1850 to 0.29 in 1850-1900. Hence, not only does \textit{Household Economics} become the dominant topic in French drama as the 19th century progresses, but it also responds strongly to GDP shocks after 1820. This result may be due to society, and playwrights in particular, placing increasingly more importance on the economy as capitalist relations deepened.

\begin{figure}[tbp]
\centering
\includegraphics[scale=0.85]{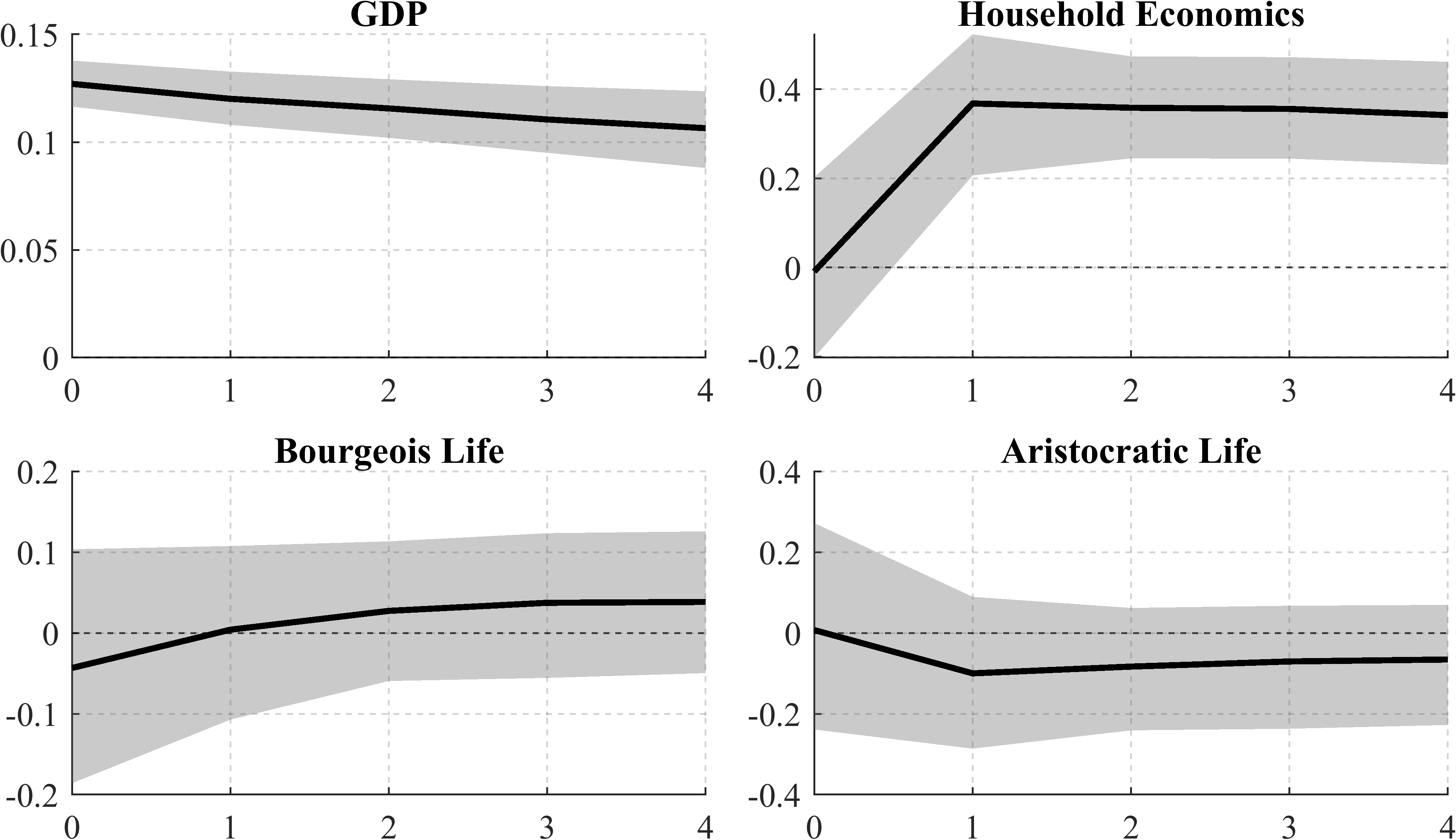}
	\caption{BVAR model, 1820–1900, GDP shock and responses of the variables}
    \caption*{\footnotesize Note: Impulse responses to a one standard deviation structural GDP shock. Solid lines denote posterior mean estimates and shaded areas denote 68\% confidence intervals. Responses whose confidence intervals exclude zero are statistically significant.}
\end{figure}

Table 3 shows the impact of the shock on the variables of interest. Notice that the values of \textit{Household Economics} are considerably higher than those of \textit{Bourgeois Life} and \textit{Aristocratic Life}, both in the short and in the long terms, denoting the greater sensitivity of that topic to GDP shocks. As a robustness check, we have extended the base model by adding human capital and life expectancy. Our results do not change when these control variables are included; they are reported in Section A2.2 of the Appendix.

\begin{table}[tbp]
\centering
\caption{Impact of the structural shock on variables, 1820–1900}
\resizebox{\textwidth}{!}{
\begin{tabular}{lcccc}
\hline
 & GDP & Household Economics & Bourgeois Life & Aristocratic Life \\
\hline
\textbf{Short term (0--4 periods)}  & 99.8 & 21.1 & 3.5 & 3.3 \\
                                   & [99.5, 99.9] & [11.1, 31.2] & [0.8, 8.1] & [0.9, 8.2] \\
\addlinespace
\textbf{Long term (20+ periods)}   & 100.0 & 95.3 & 47.5 & 47.6 \\
                                   & [99.8, 100.0] & [80.1, 99.2] & [4.9, 90.1] & [6.1, 90.6] \\
\hline
\end{tabular}
}
\caption*{\footnotesize Notes: Entries report the share of forecast error variance in each variable explained by the structural GDP shock, expressed as percentages. Short and long term correspond to horizons of 0 to 4 and 20 years respectively. Brackets denote 68\% confidence intervals. Higher values indicate greater sensitivity to economic conditions.}
\end{table}

Overall, our findings indicate that while both \textit{Bourgeois Life} and \textit{Household Economics} responded to GDP shocks, they did so in different periods. The former is sensitive to GDP shocks in the period leading to the French Revolution. The latter, on the other hand, responds to GDP shocks only once France effectively begins its industrialisation process in the 1820s. The increased sensitivity of \textit{Household Economics} to economic performance may reflect the structural transformation of French society, shifting from an agrarian to a more industrialised system. As the management of money within households gained increasing importance, dramaturgs incorporated these societal transformations into their writings.

\section{A Theoretical Model of Literary Change}

In the previous section, we showed that during the 18th century, increases in GDP were associated with a statistically significant rise only in the topic \textit{Bourgeois Life}. After the Revolution and the acceleration of French industrialisation during the 19th century, further increases in GDP instead led to a larger share of writings on \textit{Household Economics}. The present section develops a stylised model to provide a rationale for the mechanisms underlying these patterns. Building on the notion that, under fundamental uncertainty, agents rely on conventions and the observed behaviour of others to guide their decisions, we adopt a modelling structure from the discrete choice literature.\footnote{For a pedagogical presentation and literature review, see \cite{FrankeWesterhoff2017}. Seminal contributions include \cite{Kirman1993} and \cite{BrockDurlauf2001}.} Our narrative emphasises two main transmission channels. First, agents look at the choices of their peers when deciding what to write about. Second, they respond to the material conditions of their time, proxied here by the level of economic activity.

Discrete choice models have been particularly influential in behavioural macroeconomics. Boundedly rational agents interact and generate emergent dynamics that cannot be reduced to representative agent benchmarks (\citealp{Hommes2021}). Recent applications include real-financial market interactions (\citealp{BrockHommes1997}; \citealp{DieciHe2018}; \citealp{DieciSchmittWesterhoff2018}), exchange rate market dynamics (\citealp{DeGrauweGrimaldi2006}; \citealp{Proano2011}; \citealp{GardiniRadiSchmittSushkoWesterhoff2022}), monetary/fiscal policy under optimistic and pessimistic waves (\citealp{DeGrauweJi2019}; \citealp{DeGrauweForesti2023}), climate change-related processes (\citealp{Zeppini2015}; \citealp{Davila-FernandezProanoSordi2026}), and political polarisation (\citealp{LeonardLipsitzBizyaevaFranciLelkes2021}). A key feature of this class of models is the smooth nonlinear law of motion, typically represented by a hyperbolic tangent function, whose S-shaped form has a direct parallel with models of cultural transmission and socialisation (\citealp{BisinVerdier2001}; \citealp{BisinVerdier2011}). Discrete choice models have also been used for the valuation of cultural heritage (\citealp{Willis2014}; \citealp{ThrosbyZednikArana2021}) and demand for cultural goods (\citealp{LevyGarbouaMontmarquette1996}; \citealp{ThrosbySevernPetetskaya2024}). To the best of our knowledge, however, we are the first to apply this framework to the supply of drama production in France during a critical period of long-run economic and social transformation.

Suppose the universe of writers ($W$) in France at the beginning of the 18th century was constant and divided between those already engaging in \textit{Bourgeois Life} and \textit{Household Economics} topics ($W_{BL.HE}$) versus authors writing on all other possible themes ($W_{Other}$). \textit{Bourgeois Life} and \textit{Household Economics} are treated together for two reasons. First, our econometric analysis shows that both topics responded to shocks in economic activity between 1700 and 1900. Second, this allows us to keep the model very transparent, enabling a more precise identification of the channels generating each outcome. In mathematical terms, we have:%
\[
W=W_{BL.HE_{t}}+W_{Other_{t}}%
\]
Define the difference between the two groups as:%
\[
w_{t}=W_{BL.HE_{t}}-W_{Other_{t}}%
\]
so that we can introduce an auxiliary variable $x\in\left(  -1,1\right)  $
that captures the composition of the writers set at any moment in time:%
\begin{equation}
x_{t}=\frac{W_{BL.HE_{t}}-W_{Other_{t}}}{W_{BL.HE_{t}}+W_{Other_{t}}}=\frac{W_{BL.HE_{t}}}%
{W}-\frac{W_{Other_{t}}}{W}\label{x_definition}%
\end{equation}
If all authors were writing about \textit{Bourgeois Life} or \textit{Household Economics}, $W_{BL.HE_{t}}/W=1$ and
$W_{Other_{t}}/W=0$, so that $x_{t}=1$. Alternatively, if all literacy
production were concentrated on the other topics, then $W_{BL.HE_{t}}/W=0$ and
$W_{Other_{t}}/W=1$, so that $x_{t}=-1$. It is easy to see that an equal balance would imply $x_{t}=0$.

Each year, authors face the binary choice of continuing to write on their current topic or switching to a different one. This means that their shares evolve over time according to:
\begin{align}
\frac{W_{BL.HE_{t}}}{W}  & =\alpha\frac{W_{BL.HE_{t-1}}}{W}+\left(  1-\alpha\right)
p_{BL.HE_{t-1}}\nonumber\\
& \label{shares}\\
\frac{W_{Other_{t}}}{W}  & =\alpha\frac{W_{Other_{t-1}}}{W}+\left(  1-\alpha\right)
p_{Other_{t-1}}\nonumber
\end{align}
where $\alpha\in\left(  0,1\right)  $ is an inertia coefficient capturing the
share of authors that do not change topics, while $p_{BL.HE}$ and $p_{Other}$ are the probabilities of producing $BL.HE$ or $Other$ material. When $\alpha=1$, the authors' composition is constant, while for $\alpha=0$, all agents face the uncertainty of belonging to either group.

Substituting the probability functions defined in Appendix A3 and the definition of $x$ in Eq. (\ref{x_definition}) into Eq. (\ref{shares}), we obtain, after some algebraic manipulations, the law of motion governing the evolution of the writer population:

\begin{equation}
    x_{t}=\alpha x_{t-1}+\left(  1-\alpha\right)  \tanh\left(  \beta
x_{t-1}+\gamma y_{t-1}+\varepsilon_{t}\right) \label{dynamicmap}
\end{equation}
Eq. ($\ref{dynamicmap}$) captures the idea that literary production is embedded in political-economic relations, yet not fully determined by material conditions, as agents interact with one another. That is, as economic activity ($y$) accelerates and material relations are transformed, dramaturgs become more likely to address themes associated with bourgeois everyday life and the household economy. Such channel is mediated by the parameter $\gamma>0$, which stands for the importance agents place on the economy when making their choices. Playwrights, nonetheless, are also influenced by their social milieu. Hence, a writer surrounded by peers focusing on plays centred on aristocratic themes may follow similar steps, potentially countervailing the effect of accelerating economic growth. Conversely, writers connected to peers focusing on bourgeois social life and the household economy are more likely to write on similar themes, since their social setting and improved economic conditions reinforce one another. Parameter $\beta>0$ stands for the strength of this peer- or group-effect. Finally, $\varepsilon \sim \mathcal{N}(0,\sigma)$ is an i.i.d. Gaussian with zero mean error term.

\begin{figure}[tbp]
\centering
\includegraphics[scale=0.65]{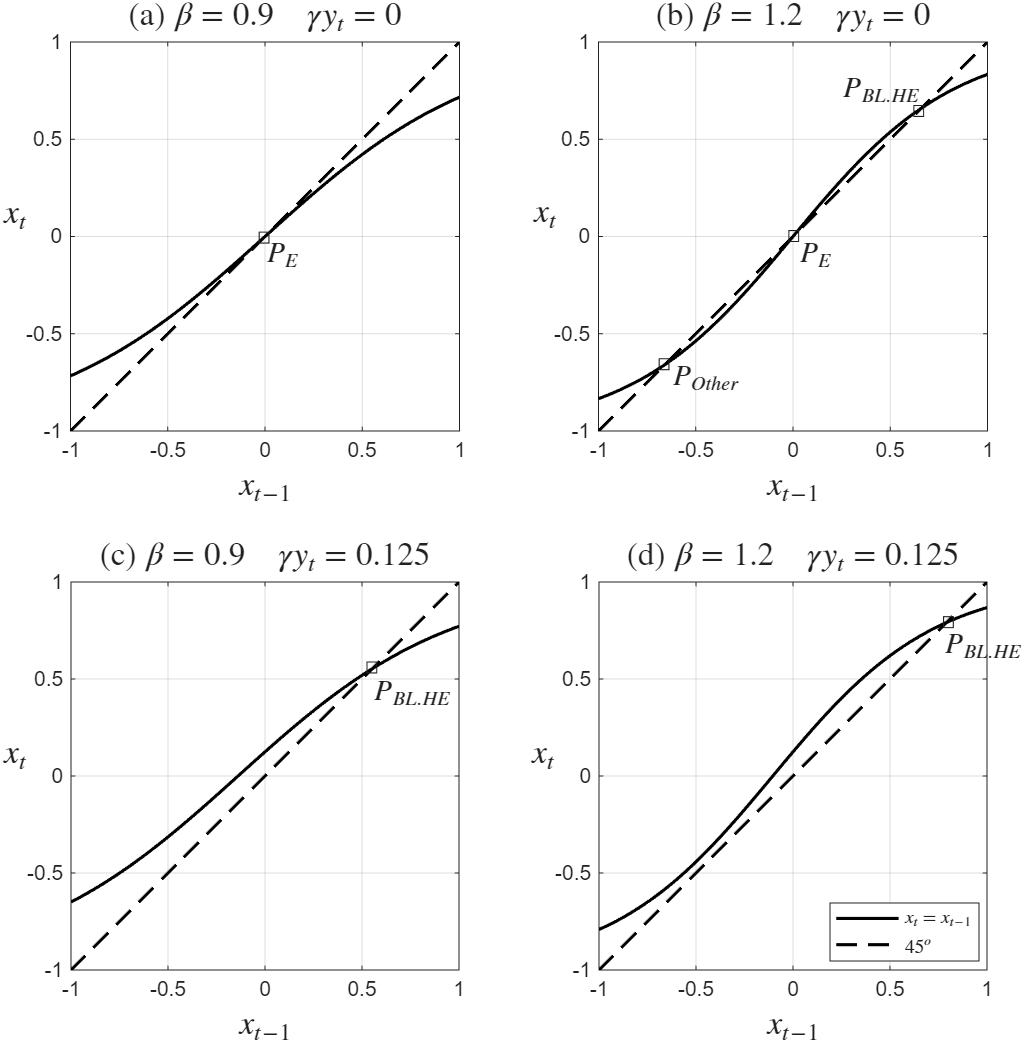}
	\caption{Unique and multiple equilibria cases under different parametrisations.}
    \label{Equilibria}
\end{figure}

In stead-state, $x_t=x_{t-1}$, and if we disregard for a moment the stochastic component, this dynamic equation admits up to three equilibrium solutions. Fig. \ref{Equilibria} illustrates four different scenarios depending on combinations of $\beta$ and $\gamma$. Panel (a) shows that when the peer effect is sufficiently small, and writers do not consider economic performance, there is polarisation between $BL.HE$ and $Other$ topics. In this case, authors are equally divided between the two groups. However, as we increase the strength of group influence, indicated in panel (b), two additional equilibria emerge: one in which \textit{Bourgeois Life} and \textit{Household Economics} dominate, and another characterised by authors who mainly write about other themes. While the peer effect creates consensus, panels (c) and (d) show that a sufficiently strong economic growth channel can effectively kill the $Other$ solution, leading authors to converge on $BL.HE$. 

\begin{figure}[tbp]
\centering
\includegraphics[scale=0.45]{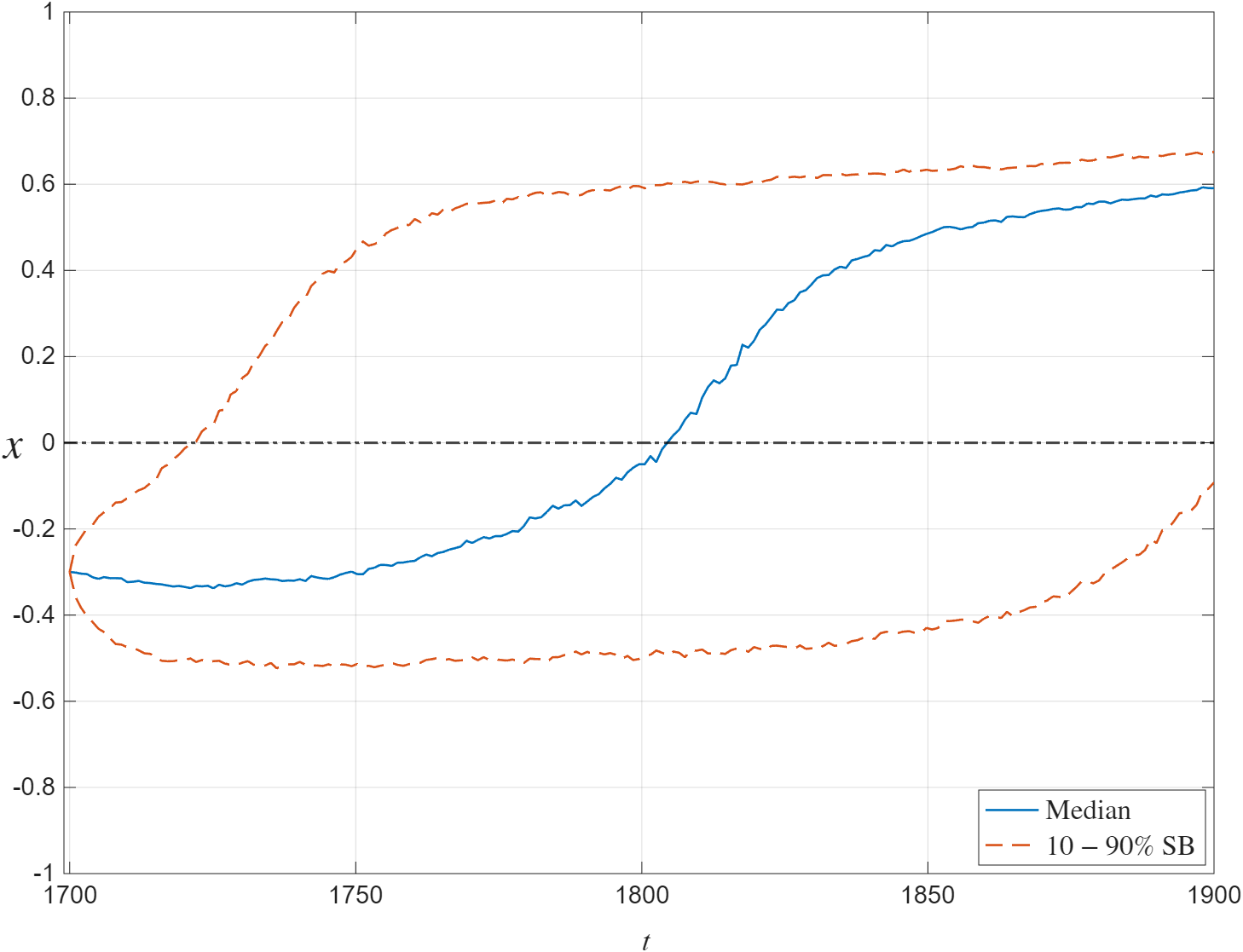}
	\caption{Monte Carlo dynamics based on 1000 independent simulations. The blue line reports the median path across repetitions, while the dashed orange lines indicate the 10-90 per cent simulation band (SB). Model parameter values are $\alpha = 0.5$, $\beta=1.11$, $\gamma=0.00001$, $\sigma = 0.1$.}
    \label{Montecarlo}
\end{figure}

Reintroducing stochasticity, we implement a Monte Carlo simulation over 200 periods, corresponding to the years 1700–1900, and repeat it 1000 times to capture the stochastic variability of outcomes. For each iteration, the model is initialised with a starting state, and a series of random shocks is applied to generate the evolution of the state over time. The results from all iterations are aggregated and reported in Fig. \ref{Montecarlo}. The median trajectory, in blue, as well as the 10th and 90th percentiles, in orange, provide a measure of uncertainty around the central trend. Peer-effects are responsible for the relative stability of traditional plays during the 18th century, but the balance switches to favour \textit{Bourgeois Life} and \textit{Household Economics} as the economy grows.

\begin{figure}[tbp]
\centering
\includegraphics[scale=0.575]{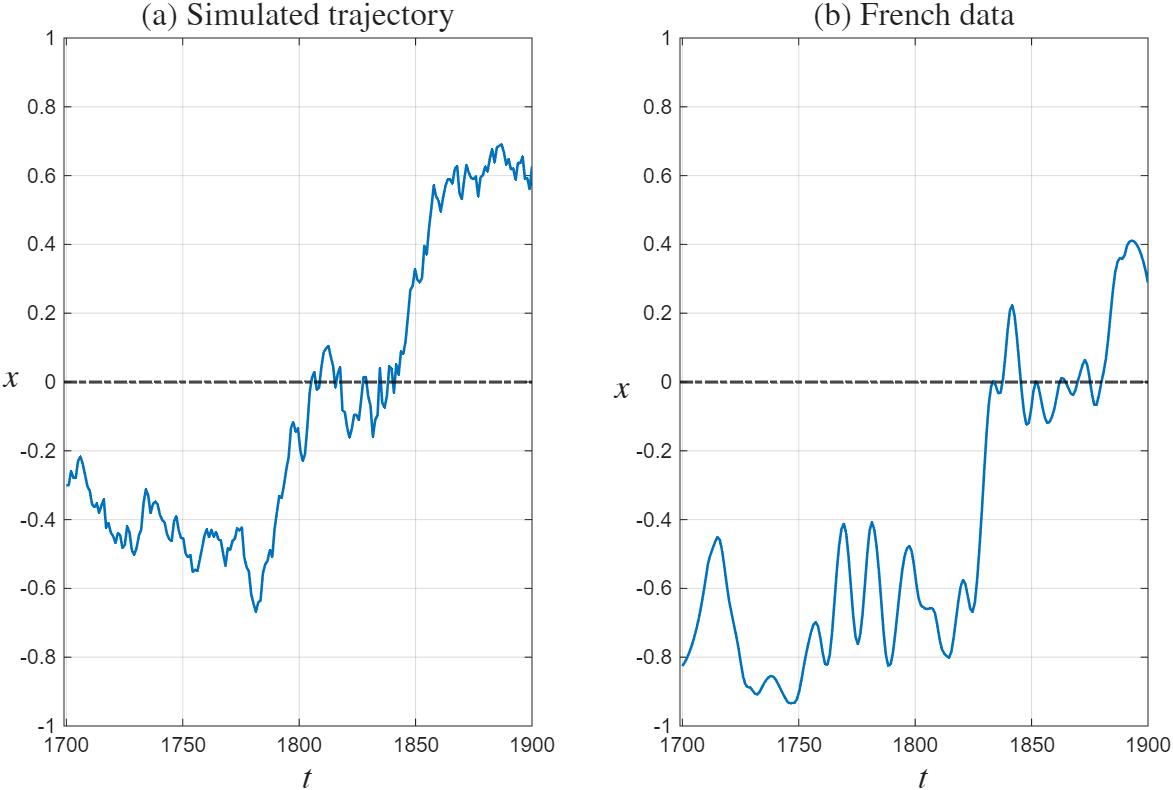}
	\caption{Comparing a simulated single trajectory with the French experience. Panel (a) uses as model parameter values $\alpha = 0.5$, $\beta=1.11$, $\gamma=0.00001$, $\sigma = 0.1$. Panel (b) fills missing values using linear interpolation with nearest-endpoint values to produce a continuous series. They were previously smoothed using a Lowess filter.}
    \label{Trajectory}
\end{figure}

Fig. \ref{Trajectory} compares a single simulated trajectory of the model with the variable $x$ calculated from our dataset. The simulated and empirical paths share the same overall trend and broadly similar fluctuation patterns. They differ, however, in the precise timing of the main transitions and in the value of $x$ following each jump. This is expected given the stylised nature of the model, which abstracts from the many historical contingencies. The exercise's purpose is not to replicate the historical record but to show that peer effects and economic conditions are jointly sufficient to generate dynamics of the observed qualitative character. This claim is supported more rigorously by the Monte Carlo simulation bands reported in Fig. 6.

\section{Conclusion}

The political-economic transformations of the eighteenth and nineteenth centuries left measurable traces in French literary production. Applying LDA and JSD to a corpus of 1,215 French plays, we have shown that the topical distribution of French drama changed profoundly between the late eighteenth and the mid-nineteenth centuries. The topic \textit{Aristocratic Life}, emphasising sovereignty, virtue, law, and political authority, declined steadily in prevalence over time, while \textit{Bourgeois Life} and \textit{Household Economics} gained prominence after the French Revolution.

Our main claim is that the thematic profile of French drama radically changed as the absolutist state weakened and capitalist social relations developed. Themes centred on political authority and aristocratic values gave way to more mundane concerns centred on everyday life, domestic relations, and household finances. BVAR models under max-share shock identification reveal that this transformation followed a precise two-stage dynamic: \textit{Bourgeois Life} responded to GDP shocks in the eighteenth century, while \textit{Household Economics} became sensitive to economic performance only after 1820, tracking the acceleration of French industrialisation. A stylised discrete-choice model shows that these dynamics could emerge from two interacting mechanisms, namely peer effects among authors and sensitivity to prevailing economic conditions. Monte Carlo simulations reproduce the observed historical trajectory with reasonable fidelity.

These findings carry implications that extend well beyond the French case. They suggest that cultural production responds to economic transformation through identifiable social mechanisms in which collective behaviour and material conditions interact. The two-stage dynamic we document, with different cultural registers responding to economic growth at different moments, indicates that capitalist development reshapes not only material life but also the symbolic universe through which societies interpret and give form to human life. The question of how economic shocks propagate into cultural discourse, and through what mechanisms, remains largely open. By showing that computational linguistics, econometric methods, and formal economic modelling can be combined to trace these processes at scale and over the long run, this paper offers a framework that can be extended to other national traditions, other literary genres, and other periods of structural economic change.

\section*{Acknowledgments} 

Thiago Dumont Oliveira would like to acknowledge that he has received funding from the European Union’s Horizon Europe research and innovation programme under grant agreement No 101186601.

\bibliographystyle{plainnat}
\bibliography{references}         


\newpage

\appendix

\numberwithin{equation}{section}
\numberwithin{figure}{section}
\numberwithin{table}{section}

\section{Appendix} 

\subsection{Non-Negative Matrix Factorisation} 


As a robustness check to the LDA analysis, we use Non-Negative Matrix Factorisation (NMF), a deterministic linear-algebraic method widely used in topic modelling (\citealp{lee1999learning}, \citealp{lee2000algorithms}). Table A.1 presents the 10 topics obtained using NMF. While there are some interesting differences with respect to LDA, this alternative topic modelling approach yields results similar to those obtained using LDA. 

Topic 3, obtained through NMF, seems to combine topics \textit{Bourgeois Life} and \textit{Household Economics}. 12 out of its 15 top words overlap with the top words of \textit{Household Economics}: \textit{poor}, \textit{mother}, \textit{father}, \textit{money}, \textit{husband}, \textit{franc}, \textit{devil}, \textit{eat}, \textit{house}, \textit{letter}, \textit{affair}, \textit{boy}. Yet, it also contains four words in common with the topic \textit{Bourgeois Life}, i.e. \textit{poor}, \textit{mother}, \textit{old}, \textit{evening}. Moreover, notice that \textit{poor} is the top word of both topic 3, in Table A.1, and of \textit{Bourgeois Life}, in Table 1. \textit{Mother}, the second most important word of topic 3, does not figure among the top 15 words of \textit{Household Economics}, which further suggests that it combines \textit{Bourgeois Life} and \textit{Household Economics} into the same topic. 

This result reinforces our decision to aggregate these two variables in the modelling exercise; while there are differences between these two topics, one emphasising economic aspects, the other everyday social life themes, both deal with the emerging middle class and are fundamentally distinct from aristocratic and courtly life themes that had prevailed in French drama hitherto.  The average prevalence of the topic 3 increased from 0.16 in the 18th century to 0.41 in the 19th century, which is roughly the same variation as \textit{Bourgeois Life} and \textit{Household Economics}, when considered together, which increased from 0.15 to 0.45 in the same period, as reported in Table 1. 

The other topic of interest, \textit{Aristocratic Life}, gets split into 3 topics. Apart from topic 1 on Table A.1, which overlaps considerably with \textit{Aristocratic Life}, topics 8 and 9 also clearly point to courtly life themes. This fact is indicated by the presence of words such as  \textit{knight}, \textit{viscount}, \textit{milord}, \textit{opera} and \textit{countess} in topic 8; and \textit{king}, \textit{prince}, \textit{queen}, \textit{princess}, \textit{throne}, \textit{court}, \textit{majesty} and\textit{ crown} in topic 9. Taken together, the average prevalence of these three topics falls from 0.27 in 1700-1800 to 0.12 in 1800-1900. Such a result is quite similar to that previously obtained for the topic \textit{Aristocratic Life} using LDA, whose prevalence fell from 0.25 to 0.09 over the same period.  

\begin{table}[tbp]
\centering
\begingroup
\scriptsize
\setlength{\tabcolsep}{3pt}
\renewcommand{\arraystretch}{1.05}
\sloppy
\begin{tabularx}{\textwidth}{@{}
>{\centering\arraybackslash}p{0.7cm}
>{\centering\arraybackslash}p{2.2cm}
>{\centering\arraybackslash}p{2.2cm}
>{\RaggedRight\arraybackslash}p{3.4cm}
>{\RaggedRight\arraybackslash}X@{}}
Topic & \shortstack{Prevalence$_{1700-1800}$}
      & \shortstack{Prevalence$_{1800-1900}$}
      & Top Words (15)
      & Representative Works \\
\midrule

0 & 0.15 & 0.09 &
marquis, uncle, Julie, countess, knight, sister, Sophie, brother, affair, baron, letter, marriage, niece, marry, sentiment &
Carmontelle: \textit{La Cloison, Beaucoup de Peine pour Rien}\newline
Carmontelle: \textit{Les Moeurs du Temps}\newline
Carmontelle: \textit{La Cloison, Comédie}\newline
Carmontelle: \textit{Condamner sans Entendre, ou La Statue}\newline
Carmontelle: \textit{La Statue} \\
\addlinespace[4pt]

1 & 0.15 & 0.05 &
blood, son, death, fury, crime, people, fate, horror, cruel, law, virtue, tyrant, glory, again, destiny &
Racine: \textit{Polixène}\newline
Voltaire: \textit{Zulime}\newline
Racine: \textit{Bajazet}\newline
Guimond de La Touche: \textit{Iphigénie en Tauride}\newline
Lebrun: \textit{Les Druides} \\
\addlinespace[4pt]

2 & 0.06 & 0.12 &
Molière, theatre, author, play, actor, comedy, poet, stage, talent, comedian, perform, art, role, genius, laughter &
Molière: \textit{Mascarille}\newline
Sedaine: \textit{Le Quinze Janvier}\newline
Collé: \textit{L'Assemblée}\newline
Chalmeton: \textit{A Jean Racine}\newline
Bernard: \textit{Le Malade Réel} \\
\addlinespace[4pt]

3 & 0.16 & 0.41 &
poor, mother, father, money, husband, old, franc, devil, wine, eat, evening, house, letter, affair, boy &
Labiche: \textit{La Rose Rouge}\newline
Anicet-Bourgeois: \textit{Les Aventures de Ninette}\newline
Courteline: \textit{Le Nez du Général Suif}\newline
Léonce: \textit{Le Lapin de la Portière}\newline
Richepin: \textit{Cochon de Cocher !} \\
\addlinespace[4pt]

4 & 0.17 & 0.08 &
lover, tender, ardor, kind, sweet, object, fire, beauty, flame, care, husband, charm, vow, goddess, Leander &
Hénault: \textit{Le Temple des Chimères}\newline
Saint-Gilles Lenfant: \textit{La Feinte Heureuse}\newline
Favart: \textit{L'Amant Déguisé}\newline
Vadé: \textit{La Noce de Village}\newline
Rousseau: \textit{Pan et Doris, Pastorale} \\
\addlinespace[4pt]

5 & 0.04 & 0.02 &
Rome, Caesar, senate, Roman, Romans, Cato, Brutus, Sulla, tyrant, people, Sextus, virtue, Pompey, blood, Tullia &
Voltaire: \textit{La Mort de César}\newline
Voltaire: \textit{Rome Sauvée}\newline
Voltaire: \textit{La Mort de César}\newline
Crébillon: \textit{Catilina}\newline
Deschamps: \textit{La Mort de Caton} \\
\addlinespace[4pt]

6 & 0.10 & 0.02 &
Lisette, Angélique, Valère, Bian, Dorante, Frontin, marry, Damis, marriage, Lucile, mother, mistress, affair, Dorante, Argante &
Regnard: \textit{Crispin, rival de son maître}\newline
Regnard: \textit{Le Neveu de la Marquise}\newline
Marivaux: \textit{L'Épreuve}\newline
Marivaux: \textit{Le Dénouement Imprévu}\newline
Marivaux: \textit{La Mère Confidente} \\
\addlinespace[4pt]

7 & 0.07 & 0.14 &
son, mother, death, Zeus, earth, dwelling, city, kill, brother, speech, Orestes, foreigner, old man, sister, murder &
Sophocle (tr. Leconte de Lisle): \textit{Philoctète}\newline
Legouvé: \textit{Agar dans le Désert}\newline
Sophocle (tr. Leconte de Lisle): \textit{Les Trachiniennes}\newline
Eschyle (tr. Leconte de Lisle): \textit{Électre}\newline
Eschyle (tr. Leconte de Lisle): \textit{Les Choéphores} \\
\addlinespace[4pt]

8 & 0.06 & 0.02 &
abbot, knight, president, supper, viscount, play, milord, dinner, horse, opera, dog, countess, card, Bernard, tragedy &
Collé: \textit{L'Abbé de Coure-Dîner}\newline
Carmontelle: \textit{Les Voyageurs}\newline
Collé: \textit{L'Après-Dinée}\newline
Carmontelle: \textit{L'Histoire}\newline
Carmontelle: \textit{Le Portrait} \\
\addlinespace[4pt]

9 & 0.06 & 0.05 &
king, prince, queen, princess, people, state, throne, court, minister, sire, majesty, France, soldier, war, crown &
Collé: \textit{L'Officier du Gobelet}\newline
Collé: \textit{Roi et Reine}\newline
Regnard: \textit{La Méprise}\newline
Regnard: \textit{La Comédie}\newline
Vadé: \textit{Bois-Rosé} \\

\end{tabularx}
\caption{\small Topic Prevalence 1700--1800 vs. 1800--1900, 15 Top Words and 5 Representative Works.}
\endgroup
\end{table}

\subsection{BVAR Analysis: Model Specification \& Robustness Checks}

\subsubsection{BVAR Model}

The basic structure of our VAR model is given by:
\begin{equation}
Y_t=\sum_{i=1}^{p} C_i Y_{t-i}+u_t
\label{eq:var}
\end{equation}
where  $Y_t$ is a vector of variables, $C_i$ are the coefficient matrices, $Y_{t-i}$ indicates lagged endogenous variables, and $u_t$ is the vector of innovations i.e. reduced-form shocks.

The goal in shock identification is to obtain a structural vector, $\varepsilon_t$, that allows $u_t$ to be written as:

\begin{equation}
u_t=A(L)\varepsilon_t
\label{eq:structural}
\end{equation}
where
\begin{equation*}
    A(L)\equiv\sum_{k=-\infty}^{\infty}A_kL^k
\end{equation*}
is a lag polynomial in which past and future values define the vector of innovations.

In  SVAR models, $A(L)$ can be identified using the Cholesky ordering, which transforms $A(L)$ into a contemporaneous impact matrix:

\begin{equation}
A(L)=A_0
\label{eq:cholesky}
\end{equation}

From Equation \eqref{eq:cholesky}, the vector of innovations is written as:

\begin{equation}
u_t=A_0\varepsilon_t
\label{eq:impact}
\end{equation}

We can include the same variables used in the BVAR model presented in the paper $(gdp_t, household_t, bourgeois_t, aristocratic_t)^\prime$. The targeting of the shock on GDP ($gdp_t$) is defined as:
\begin{equation}
gdp_t=e_m'Y_t
\label{eq:target}
\end{equation}
such that $e_m=[1,0,0,0]'$.

The moving-average representation of Equation \eqref{eq:var} is:
\begin{equation}
Y_t=\sum_{k=0}^{\infty}B_kA_0\varepsilon_{t-k}
\label{eq:ma}
\end{equation}
where $B_k$ is the coefficient of the infinite moving-average representation and $B(L)\equiv(I-C(L))^{-1}$.

Moreover, the forecast error variance decomposition (FEVD) at horizon $h$ for the target variable is:
\begin{equation}
gdp_{t+h}-E_{t-1}(gdp_{t+h})=
e_m'
\left[
\sum_{\tau=0}^{h-1}
B_\tau A_0 \varepsilon_{t+h-\tau}
\right]
\label{eq:fevd}
\end{equation}
Thus, the contribution of the first element of the structural vector $\varepsilon_t$ is:
\begin{equation}
\mathrm{Var}
\left(
gdp_{t+h}
-
E_{t-1}(gdp_{t+h})
\,\big|\,
\varepsilon_{jt}=0,\; j=2,\ldots,4
\right)
=
e_m'
\left[
\sum_{\tau=0}^{h-1}
B_\tau A_0 e_1 e_1' A_0' B_\tau'
\right]
e_m .
\label{eq:maxshare}
\end{equation}

In Equation \eqref{eq:maxshare}, $e_1$ is the selection vector. Consequently, $A_0e_1$ selects the first column of $A_0$. The maximisation is performed over orthogonal rotations of the structural shocks. Therefore, the max-share approach consists of choosing an orthogonal rotation matrix that maximises Equation \eqref{eq:maxshare}. This optimisation process generates $\varepsilon_{1t}$, which provides the largest contribution to the fluctuations in GDP.

\subsubsection{BVAR Analysis: Robustness Checks}

This section explores the robustness of our findings in Section 4 by extending the base model. We assess whether the results are driven by improvements in human capital and health, since this period observed progress in these areas. Thus, two variables are included. The first is human capital, measured by public education expenditure as a share of GDP. Data were collected from the World Human Capital Expenditure Database (\url{https://whce.world/data/}). The second variable is life expectancy, defined as the average number of years a person lives. Time series were retrieved from the Our World in Data (\url{https://ourworldindata.org/search}).

The max-share shock identification implements a structural GDP shock considering the household, aristocratic, and bourgeois indices, with the addition of human capital and health. 
Given that human capital and health are available only in the second period, Fig. A.1 reports impulse responses for a structural GDP shock in the period 1820–1900. Human capital and health increase due to the shock, but the estimates are not statistically significant. We still document an increase in household economics, consistent with the results from the base model. 

\begin{figure}[tbp]
\centering
\includegraphics[scale=0.9]{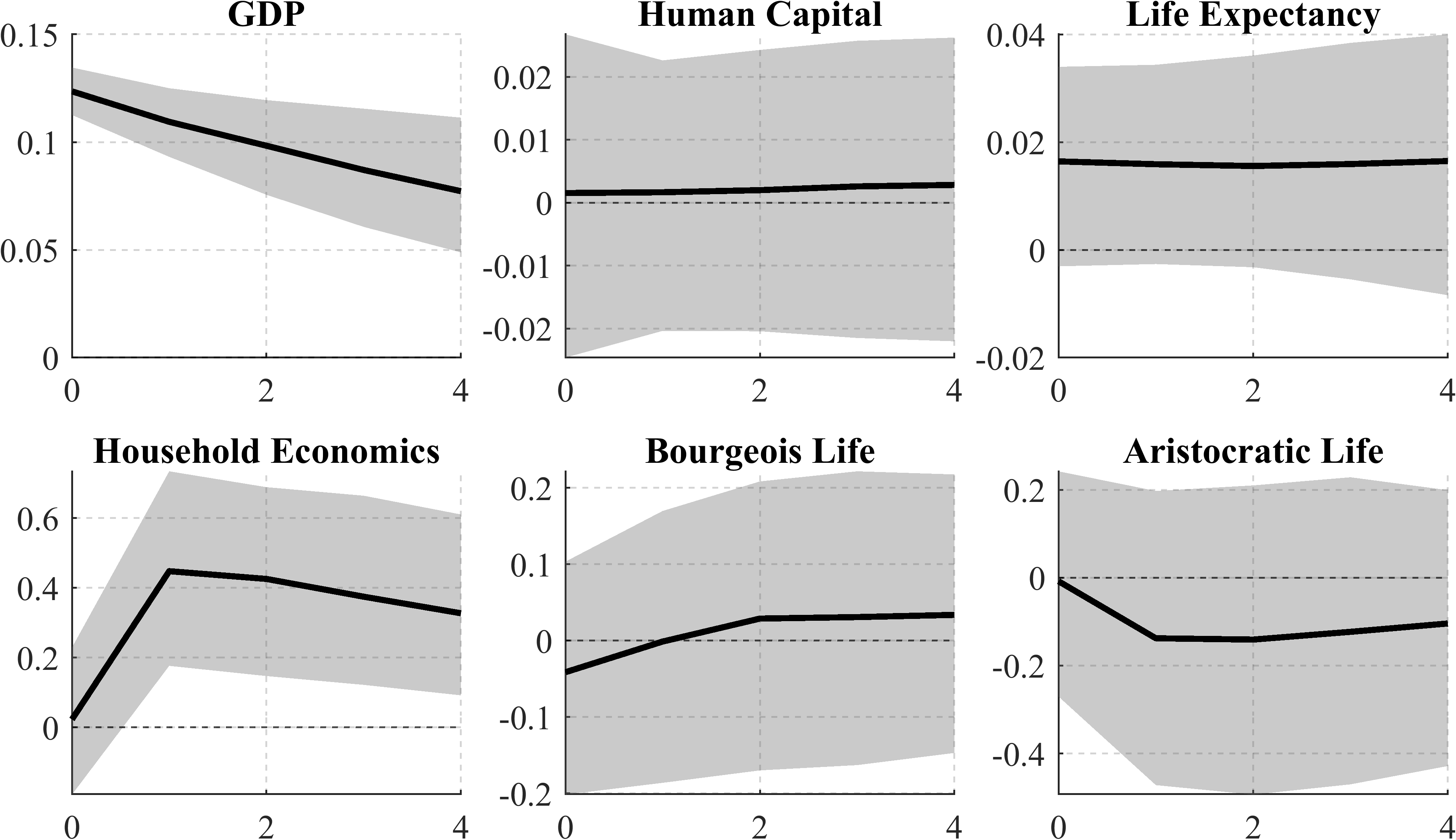}
	\caption{BVAR model, 1820–1900, GDP shock and responses of the variables}
    \caption*{\footnotesize Note: Impulse responses to a one standard deviation structural GDP shock. Solid lines denote posterior mean estimates and shaded areas denote 68\% confidence intervals. Responses whose confidence intervals exclude zero are statistically significant.}
\end{figure}

Table A.2 replicates Table 3 with our additional control variables. One qualitative improvement is that the GDP column shows that the structural shock does not explain 100\% of GDP fluctuations. Although the shock analysis in Fig. A.1 shows that human capital and health are not sensitive to the GDP shock, Table A.2 indicates that this shock explains 13.8\% of the variation in human capital in the last period and 11\% of the variation in health in the last period. As previously observed, \textit{Household Economics} is the most sensitive index to the structural GDP shock. Thus, incorporating public and health dimensions does not change our main results.

\begin{table}[tbp]
\centering
\caption{Impact of the structural shock on variables, 1820–1900}
\resizebox{\textwidth}{!}{
\begin{tabular}{lcccccc}
\hline
 & GDP & Human Capital & Life Expectancy & Household Economics & Bourgeois Life & Aristocratic Life \\
\hline
\textbf{Short term (0--4 periods)}  & 90.1 & 6.7 & 5.2 & 22.9 & 7.6 & 8.6 \\
                                   & [78.4, 96.8] & [1.8, 16.7] & [1.4, 12.9] & [6.4, 42.9] & [2.2, 20.0] & [2.1, 21.1] \\
\addlinespace
\textbf{Long term (20+ periods)}   & 30.8 & 13.8 & 11.0 & 27.9 & 18.8 & 20.2 \\
                                   & [7.5, 73.3] & [2.6, 48.6] & [1.3, 48.8] & [5.3, 65.3] & [3.1, 55.8] & [3.6, 56.1] \\
\hline
\end{tabular}
}
\caption*{\footnotesize Notes: Entries report the share of forecast error variance in each variable explained by the structural GDP shock, expressed as percentages. Short and long term correspond to horizons of 0 to 4 and 20 years respectively. Brackets denote 68\% confidence intervals. Higher values indicate greater sensitivity to economic conditions.}
\end{table}

\newpage

\subsection{Building the Model}

Our probability functions follow the functional forms adopted in the discrete-choice literature (e.g. \citealp{BrockHommes1997}; \citealp{FrankeWesterhoff2017}; \citealp{Davila-FernandezProanoSordi2026}), and depend on the success or fitness ($U_{i}$) of the two states $i\in \{BL.HE,Other\}$, so that:
\begin{align}
p_{BL.HE_{t-1}}  & =\frac{\exp\left(  U_{BL.HE_{t-1}}\right)  }{\exp\left(
U_{BL.HE_{t-1}}\right)  +\exp\left(  U_{Other_{t-1}}\right)  }\nonumber\\
& \label{probabilities}\\
p_{Other_{t-1}}  & =\frac{\exp\left(  U_{Other_{t-1}}\right)  }{\exp\left(
U_{Other_{t-1}}\right)  +\exp\left(  U_{BL.HE_{t-1}}\right)  }\nonumber
\end{align}

We assume that as economic activity ($y$) accelerates, authors became more likely to engage with \textit{Bourgeois Life} and \textit{Household Economics}. Moreover, a writer is more likely to write about topics their peers are also engaging with. This means:%
\begin{align}
U_{BL.HE_{t-1}}  & =\beta x_{t-1}+\gamma y_{t-1}+\varepsilon_{t}\nonumber\\
& \label{utilities}\\
U_{Other_{t-1}}  & =-\beta x_{t-1}-\gamma y_{t-1}-\varepsilon_{t}\nonumber
\end{align}
where $\beta>0$ is the strengh of the peer- or group-effect, $\gamma>0$ stands
for the importance agents place on the economy when making their choices, and
$\varepsilon \sim \mathcal{N}(0,\sigma)$ is an error term capturing exogenous disturbances. We assume it to be i.i.d. Gaussian with zero mean.

Substituting (\ref{utilities}) into (\ref{probabilities}) and the resulting expressions into Eq. (\ref{shares}), we obtain:%
\begin{align}
\frac{W_{BL.HE_{t}}}{W}  & =\alpha\frac{W_{BL.HE_{t-1}}}{W}+\frac{\left(
1-\alpha\right)  \exp\left(  \beta x_{t-1}+\gamma y_{t-1}+\varepsilon
_{t}\right)  }{\exp\left(  \beta x_{t-1}+\gamma y_{t-1}+\varepsilon
_{t}\right)  +\exp\left(  -\beta x_{t-1}-\gamma y_{t-1}-\varepsilon
_{t}\right)  }\nonumber\\
& \label{shares2}\\
\frac{W_{Other_{t}}}{W}  & =\alpha\frac{W_{Other_{t-1}}}{W}+\frac{\left(
1-\alpha\right)  \exp\left(  -\beta x_{t-1}-\gamma y_{t-1}-\varepsilon
_{t}\right)  }{\exp\left(  -\beta x_{t-1}-\gamma y_{t-1}-\varepsilon
_{t}\right)  +\exp\left(  \beta x_{t-1}+\gamma y_{t-1}\right)  }\nonumber
\end{align}
Subtracting the second expression in (\ref{shares2}) from the first, and making use of the definition in (\ref{x_definition}), it follows:%
\[
x_{t}=\alpha x_{t-1}+\left(  1-\alpha\right)  \tanh\left(  \beta
x_{t-1}+\gamma y_{t-1}+\varepsilon_{t}\right)
\]

\end{document}